# A New Mask R-CNN Based Method for Improved Landslide Detection

Silvia Liberata Ullo, *Senior Member, IEEE*, Amrita Mohan, Alessandro Sebastianelli, *Member, IEEE*,
Shaik Ejaz Ahamed, Basant Kumar, *Member, IEEE*, Ramji Dwivedi, *Member, IEEE*,
and G. R. Sinha, *Senior Member, IEEE*

*Abstract*—This paper presents a novel method of landslide detection by exploiting the Mask R-CNN capability of identifying an object layout by using a pixel-based segmentation, along with transfer learning used to train the proposed model. A data set of 160 elements is created containing landslide and non-landslide images. The proposed method consists of three steps: (i) augmenting training image samples to increase the volume of the training data, (ii) fine tuning with limited image samples, and (iii) performance evaluation of the algorithm in terms of precision, recall and F1 measure, on the considered landslide images, by adopting ResNet-50 and 101 as backbone models. The experimental results are quite encouraging as the proposed method achieves Precision equals to 1.00, Recall 0.93 and F1 measure 0.97, when ResNet-101 is used as backbone model, and with a low number of landslide photographs used as training samples. The proposed algorithm can be potentially useful for land use planners and policy makers of hilly areas where intermittent slope deformations necessitate landslide detection as prerequisite before planning.

## I. INTRODUCTION

Landslides or mudslides are an extensive phenomenon, resulting in huge upheavals worldwide with a great frequency [1-3]. It is a significant hydro-geological threat affecting large areas of the world, and in particular the India country, including the Western Ghats, North-eastern hill areas, Himalayan regions etc. The Northwest Himalayan regions of India, incorporating Himachal Pradesh, Jammu & Kashmir and Uttarakhand, are known for highest landslide hazard prone areas. Many heritage temples and Hindu pilgrim sites such as Badrinath, Kedarnath, and Kailash Mansarovar are situated in these upper Himalayan regions. Landslides occurring in Himalayan regions have been causing considerable damages to archaeological sites and severe loss of lives for many years. The world heritage site most vulnerable to landslips includes Changu Narayan Temple, Machu Picchu, and the Darjeeling Himalayan Railway.
Climate changes and anthropic activities have affected many archaeological sites in these regions in a severe way. Some of them are under an even bigger risk because of surrounding slope instability, which can activate landslides and compromise the place integrity. Many researchers have studied the landslide susceptibility, hill slope stability, risk management assessment mapping and remote sensing based heritage site management in the hilly regions worldwide, and in the Himalayan as well as in India [3-5].
A volume of research works across the world have studied the phenomena related to landslide for the risks related to cultural heritage sites and attempted to assess their possible impact. Moreover, an International Programme on Landslides (IPL) has been created, since March 2003 [6]. Major activities for landslide risk assessments, at three major cultural heritage sites in China (Lishan, Xian), Peru (Machu, Picchu) and Japan (Unzen volcano), were reported in [7]. Analysis for the threat posed to cultural heritage by landslide and avalanches were carried out for Upper Svaneti region in Georgia. Factors such as slope, land cover, lithology, and snow avalanches were also used to generate a susceptibility map [8]. As regards Italy, despite being a country relatively small in its extension, almost all types of natural risks, including earthquakes, landslides, volcanos, floods, and coastal erosions are present. Floods and landslides appear to be the major geological problems for Italian cultural heritage sites. An interesting study was carried out in [9] and [10], where an empirical Geographical Information System (GIS) based method was applied to characterize the environmental hazards for identified heritage sites in Italy.
India and Italy are among the countries with many archaeological sites under the risk of natural hazards such as landslides, subsidence, floods etc.
For this reason, we have carried out this joint work which presents a novel model for the automated detection of landslides from digital photographs of target hilly areas, acquired by UAVs, by using computationally efficient machine learning based methods. Moreover, the proposed model by its nature is easily exportable and applicable to other areas of interest.

In the last few years, various remote sensing data types have been used dealing with the monitoring of ground deformations, landslide assessment, post-wildfire burnt area assessment, heritage site and critical infrastructure monitoring [11-13].

Silvia Liberata Ullo and Alessandro Sebastianelli are associated with the Engineering Department of University of Sannio, Benevento, 82100, Italy (e-mail: ullo@unisannio.it, sebastianelli@unisannio.it).
Amrita Mohan and Ramji Dwivedi are associated with the GIS Cell, Motilal Nehru National Institute of Technology, Allahabad, Prayagraj, 211004, India (e-mail: er.amritacs@gmail.com, ramjid@mnnit.ac.in ).
Shaik Ejaz Ahamed and Basant Kumar are associated with the Motilal Nehru National Institute of Technology, Allahabad, Prayagraj, 211004, India (e-mail: ejazahmed7051@gmail.com, basant@mnnit.ac.in ).
G. R. Sinha is associated with Myanmar Institute of Information Technology (MIIT), Mandalay 05053, Myanmar (e-mail: gr_sinha@miit.edu.mm).

These remote sensing data include: optical data, Synthetic aperture radar (SAR) data, Light detection and ranging (LiDAR) measurements. Important literatures are available on various remote sensing modalities and the types of remote sensing data [11, 14-17].

Monitoring deformation of earth surface displacement and structures specifically during landslides can be investigated by utilizing several distinct types of methods and schemes [3,11,18] Moreover, extensive literature survey also on landslide photographs, UAV images and Digital Photogrammetry (DP) can be found in many research papers [19-22].

Over the years, many image-analysis-based landslide detection and assessment techniques were proposed for various types of remote sensing images [16, 23-25], with the landslide studies recently based on approaches like machine learning-based pixel and object-based image analysis [26-27].

Detailed literature survey on application of machine and deep learning for landslide detection can be found in [28-29]. Benefits of deep learning on natural images as well as the remote sensing images along with its limitations can be found in [30], where Convolutional Neural Networks (CNNs) have been utilized for extracting information from satellite optical images.

Yet, in order to retrieve remarkable results in landslide monitoring, machine and deep learning techniques require large amount of data. This is a major drawback because currently feasible dataset is not enough and is also unavailable in big quantity. Moreover, for landslide detection, exhaustive landslide dataset is required, which should contain images captured in distinct conditions. However, dataset collection, retrieval and annotation of these data is a key restriction during the overall landslide detection process. Additionally, it is an inconvenient and heavy task to retrieve high resolution data in the field of natural hazard management studies.

In this regard, since transfer learning and masking techniques have been efficiently used with distinct images such as drone images, optical remote sensing images, drone aerial images etc. to perform object detection and instance segmentation, as explained in [31-32], the interest to understand their effectiveness if applied to landslide detection arises spontaneously.

With respect to the state-of-the-art, this paper aims to present a new algorithm, which uses a Mask R-CNN [33], along with transfer learning to train the proposed model, in order to detect landslide occurrences with only fewer object images. The major contribution of the work presented in this manuscript is precisely the use of a pre-trained CNN model for the detection of landslides from photographs. Pre-trained models of CNNs, such as ResNet-50 or ResNet-101 chosen in our case, have demonstrated to not require large set of data for training, as the method can exploit the pre-trained learning model performance. Therefore, the proposed Mask R-CNN model can give reasonably good results using very small data set.

An augmentation technique is applied to increase the number of image data, and the concept of transfer learning is extended to landslide detection, showing to work well with limited amount of labelled data.

The main objectives of the present work are:

(a) To perform object detection task on landslide photographs for achieving landslide detection with minimal dataset using a pre-trained Mask R-CNN model
(b) To use transfer learning models for achieving good results with finite training samples.
(c) To Collect freely available landslide photographs from distinct sources.

The above objectives will be fulfilled through the following activities:

1. Augmenting training samples to increase the volume of the training data (Set A and Set B)
2. Fine tuning with limited photographs
3. Investigating pre-trained learning model performance by using ResNet-50 and ResNet-101 as backbone neural networks.

The rest part of this paper is organized as follows: Section 2 explains the algorithmic procedure for landslide detection. Section 3 provides details of the proposed method for landslide detection. Section 4 presents the simulation results. Section 5 summarizes the work and section 6 provides conclusion along with future directions.

## II. ALGORITHMIC PROCEDURE

In this section, methodology adopted for landslide detection is explained, followed by a brief introduction to transfer learning and Mask-RCNN.

(i) Transfer learning: The main aim of transfer learning [31] is to transfer knowledge from a substantial dataset (i.e. source domain) to a smaller dataset (i.e. objective domain) as shown in Fig.1. In the proposed work, transfer learning is used since it allows to deal with a smaller number of landslide images.

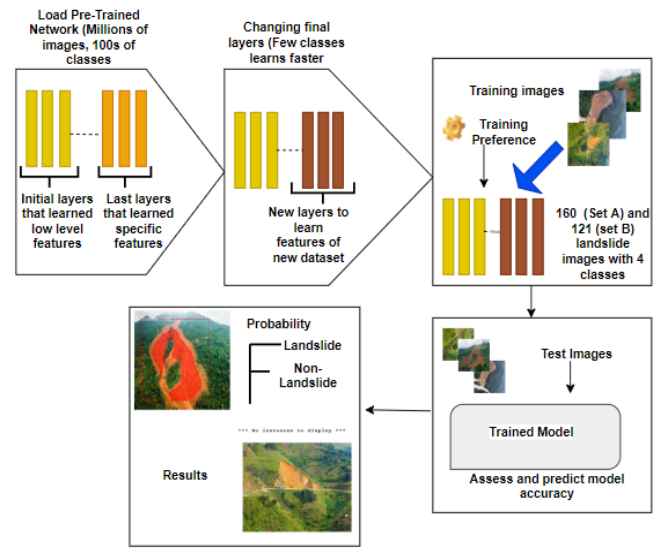

Fig. 1. Transfer Learning for landslide detection

(ii) COCO (common objects in context) Dataset [32]: In this work, trained weights of COCO dataset are utilized for landslide detection model, and the last output layer only is trained by using landslide labelled images. COCO pre-trained weights are used for both ResNet-50 and ResNet-101 CNNs.

(iii) ResNet: It is also known as Residual Network. It is a powerful framework in training a deep neural network. In our work, ResNet-50 and ResNet-101 are used. ResNet-50 is a deep residual network with 50 layers which is a subclass of the convolutional network and most popularly used model for image classification. Similarly, ResNet-101 is a convolutional neural network having 101 layers. In both CNNs (ResNet-50 and ResNet-101), bottleneck design has been used. There are five stages in ResNet architecture but in our work, we have used only 4 stages, since we are using transfer learning.

(iv) Mask R-CNN: It is an extension of Faster R-CNN to pixel-level image segmentation [33]. The core point is to decouple the pixel-level mask prediction task and the classification. Based on Faster R-CNN framework, Mask R-CNN adds a third branch for prediction of object mask in parallel with current branches to perform localization and classification.

### III. PROPOSED METHOD OF LANDSLIDE DETECTION

In our work, we have explored landslide detection method on open source landslide photographs, acquired by UAVs, using a Mask R-CNN as deep neural network, along with transfer learning for training the proposed model. The proposed landslide detection framework consist of: Data Processing (for resizing and augmentation of the images); Data Labelling (for defining the classes i.e. Landslide, Vegetation, Water body, Building and Background), and Dataset Creation (Set A and B), as shown in the Figure 2.

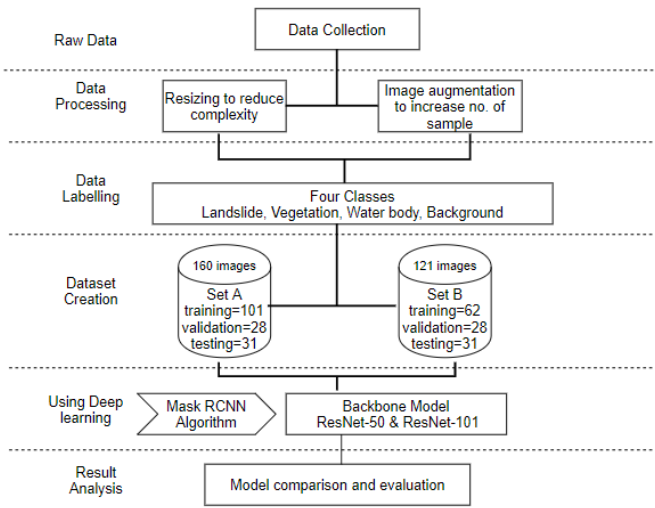

Fig. 2. Proposed landslide detection method based on Mask R-CNN and transfer learning.

The proposed algorithm comprises of: Backbone network, Region proposal network, Mask representation, and Region of Interest (RoI) Align layer. This ROI layer has been introduced as new layer for Mask R-CNN in [33] to solve misalignment problems.

In the proposed work, ResNet-the 50 and ResNet-101 are used as backbone networks for generating feature maps from landslide images. Feature maps generated from previous layers are passed through convolutional filter producing the variable size of anchor boxes containing various objects in the image. These anchor boxes are passed into parallel branches determining the objectness score of landslides and regress the bounding box coordinates. Then, a fully connected layer is used for mask prediction on landslide images, and in order to generate its input, a fully connected RoI Align is used for separating masked object from background image. The purpose of RoI Align is to convert the feature map with different sizes into a fixed-size feature map. A mask encompasses spatial information about the object, i.e. landslide in the input images.

(a) *Data Collection:* In recent years, closed-range remotely sensed images obtained from UAV (Unmanned Aerial Vehicle) photogrammetry have shown intense growth in landslide studies [23]. The dataset used in this research work is composed by several images including high resolution digital photographs, mainly acquired by UAVs, and freely downloaded from the available search engines (i.e. Bing, Google) and other sources, through a Python script written to this purpose. Some specific images that belong to particular terrains having landslides, were manually selected by visual inspection to analyze those photos that appear to be most representative for the landslide detection in the hilly region, as shown in the Figure 3.

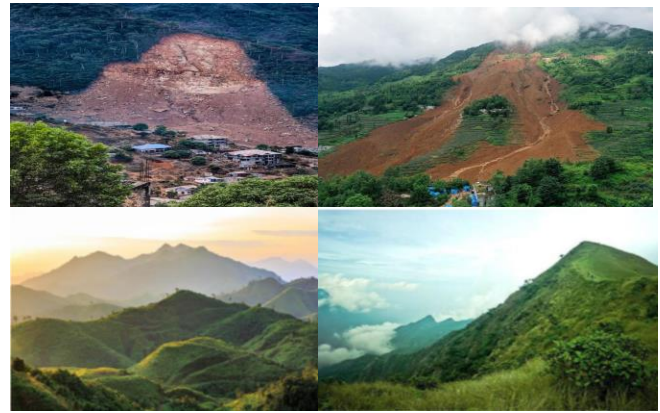

Fig. 3. Sample photographs with and without landslides.

(b) *Data Processing:* A significant contribution in this paper is the creation of a landslide image database and preprocessing of training data, since there is no photograph-based landslide imagery database available. Therefore, all downloaded landslide images are classified and assessed manually. After the selection of the images, two preprocessing steps are applied: (i) resizing the image to reduce the complexity: resized images have a size of 512*512 pixels; (ii) image augmentation applied by using the Image data generator (from the Python library) for increasing the number of training images, and improving the training suitability for the deep learning algorithm, as represented in the Figure 4.

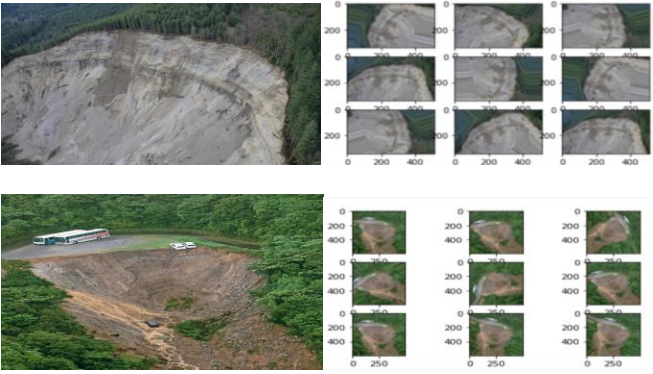

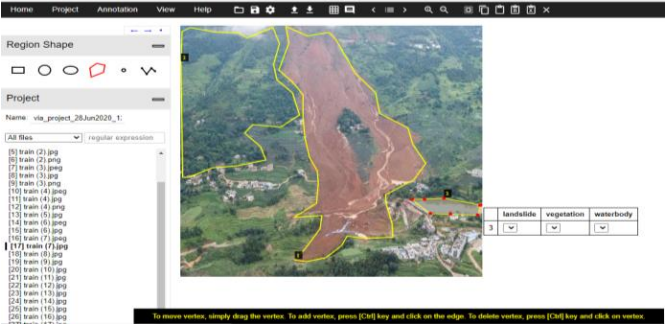

Fig. 4. Augmented outputs of sample images

(c) *Data Labelling and Annotation:* Data labelling is done to prepare the dataset for training the deep learning algorithm. Labelled data contain meaningful tags that are informative. In this work, five classes have been defined, namely: Landslide, Vegetation, Water body, Buildings, and Background, as shown in the Figure 5. For data annotation, the "VGG Image Annotator (VIA)" is used to define regions in an image and to add textual descriptions of these regions. VIA is an open-source project, based on HTML, CSS and JavaScript, and developed at the Visual Geometry Group (VGG) [34,35]

Fig. 5. Image annotation using the VGG Image Annotator (VIA) developed by the Visual Geometry Group (VGG).

(d) *Dataset Creation:* Dataset is divided into three folders: training, testing, and validation. Training data are part of the data, which help the deep learning model to make predictions. Validation data help to know whether the model is capable of correctly identifying the new examples or not, and also, they include images, used by the model to check and keep track of its learning. Testing folder contains images from which the model accuracy can be predicted. Two datasets are created Set A (Training = 101, Validation = 28, Testing = 31) and Set B (Training= 62, Validation= 28, Testing= 31). The purpose of creating two data sets of different sizes, is to observe the performance consistency of the proposed landslide detection model.

(e) *Mask R-CNN for landslide detection*: the Mask R-CNN algorithm proposed in this work, and its implementation, operates through an open-source library, to perform various tasks (TensorFlow, in this case). All experimental works are conducted on Google Colaboratory (Google Colab). Colab notebooks are capable of executing code on Google's cloud servers. Moreover, Microsoft (MS) COCO pre-trained weights for ResNet-50, and ResNet-101 models are used. As underlined before, 5 classes are used for the considered images, although MS COCO dataset makes available 80 different classes. In the Figure 6 the Mask R-CNN scheme is shown.

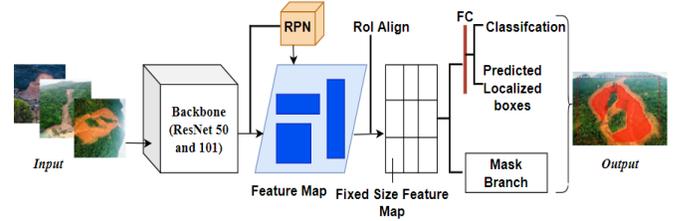

Fig. 6. Mask R-CNN for image-based landslide detection

In the proposed Mask R-CNN model, digital photographs from the created data sets are annotated and the file (.json extension) containing the coordinates of the defined classes is generated. The considered transfer learning backbone model will be trained using this file. Weights of head layers are generated, and when the testing images are passed through the considered ResNet model, the different regions are separeted using a Regional Proposal Network (RPN), helping to produce high-quality regions. In this way, different feature maps are generated corresponding to different regions.

RoI Align unit is used and a fixed size feature map is geneareted, which will be used for detecting the landslide regions on the image. Accuracy of the formed regions is determined by the trained transfer learning model.

The fixed size feature map is convolved over the image and the corresponding objectness scores are generated. These scores are indicative of the target object presence in the image under consideration, and vary between 0 to 1. Fully connected layers, using a particular objectness score threshold value, classify the image into two classes i.e. landslide and non-landslide (i.e. binary classification). Generally, a threshold of 0.8 is used in this case and a gradient descent optimization algorithm is applied to update the training parameters such as the learning rate. Learning rate for our proposed mask R-CNN model has been set to 0.001. Furthermore, the number of the epochs has been optimized by considering the obtained loss error values.

IV. SIMULATION RESULTS

In Table I the overall loss value generated after each epoch is given for ResNet-101 and 50.

TABLE I
OVERALL LOSS VALUES OF RESNET-50 AND RESNET -101
FOR LANDSLIDE DETECTION

| Epoch | ResNet-50 Loss Value | ResNet-101 Loss Value |
|---|---|---|
| 1 | 0.7370 | 0.4272 |
| 5 | 0.0689 | 0.0418 |
| 10 | 0.0384 | 0.0277 |
| 15 | 0.0329 | 0.0256 |
| 20 | 0.0286 | 0.0218 |

We have observed that the overall loss reduces while landslide detection accuracy increases for both the ResNets. Yet, it has been also observed that overtraining doe not produce benefits since it increases the loss and reduces the accuracy.

In this section, some definitions are introduced, necessary for the comprehension of the runned simulations and performance evaluation of the proposed method.

(i) *Accuracy Assessment:* Effectiveness and performance of the proposed landslide detection algorithm are evaluated using three quantitative parameters, i.e. Precision, Recall and F1 scores.

The Precision measure is used for finding the correctness of landslide area detection.

The Recall measure is used for defining how much of the actual landslide regions were identified in the image. The balance between Precision and Recall measure is calculated by using the F1 score.

The above measures can be calculated using the following equations:

$$Precision(P) = \frac{TP}{TP+FP} \quad (1)$$

$$Recall(R) = \frac{TP}{TP+FN} \quad (2)$$

$$F1\ score = 2 * \frac{P*R}{P+R} \quad (3)$$

where TP stays for True Positive, and it represents the landslide area accurately described by the applied methodology. FP stays for False Positive, and it is defined as non-landslide regions detected by the applied methodology as a landslide region in the image, whereas FN stays for False Negative and it shows actual landslide regions that are not detected by the applied methodology. The Precision, Recall and F1 scores calculated for Dataset A and Dataset B are shown below in Table II and III.

TABLE II
PRECISION, RECALL & F1 SCORE FOR DATASET A (160 IMAGES)

| Model | Type | P | R | F1 Score |
|---|---|---|---|---|
| ResNet 50 | Landslide | 1 | 0.87 | 0.93 |
| | Non-Landslide | 0.89 | 1 | 0.94 |
| ResNet 101 | Landslide | 1 | 0.93 | 0.97 |
| | Non-Landslide | 0.94 | 1 | 0.97 |

TABLE III
PRECISION, RECALL & F1 SCORE FOR DATASET A (121 IMAGES)

| Model | Type | P | R | F1 Score |
|---|---|---|---|---|
| ResNet 50 | Landslide | 1 | 0.73 | 0.85 |
| | Non-Landslide | 0.8 | 1 | 0.89 |
| *ResNet 101* | Landslide | 0.93 | 0.87 | 0.9 |
| | Non-Landslide | 0.88 | 0.94 | 0.91 |

*(ii) Frames Accuracy:* Mask R-CNN with its constitutive parts, i.e. Faster R-CNN and RPN, is used to check objectness score and detection accuracy. The proposed object detection model is tested on landslide images, and overall accuracy of the model can be observed from table IV and V. Rectangular bounding boxes are generated along with the frame if the accuracy of the detected object is above 80%. Generally, accuracy will range between 95%-100%. Accuracy, the percentage of correctly predicted data out of all the data, can be calculated as:

$$Accuracy = \frac{TP+TN}{TP+TN+FP+FN} \quad (4)$$

where, TP, FP, FN are as above defined, and TN stays for True Negative, and it represents the non-landslide area accurately described by the applied methodology.

TABLE IV
RANDOM FRAMES ACCURACY OUTPUT FOR RESNET-50

| Frame | Detection Accuracy (%) | Frame | Detection Accuracy (%) |
|---|---|---|---|
| 1 | 99 TN | 6 | 99 FN |
| 2 | 98 TN | 7 | 98 FN |
| 3 | 97 TP | 8 | 91 FN |
| 4 | 90 TP | 9 | 96 TP |
| 5 | 98 TN | 10 | 90 FN |

TABLE V
RANDOM FRAMES ACCURACY OUTPUT FOR RESNET-101

| Frame | Detection Accuracy (%) | Frame | Detection Accuracy (%) |
|---|---|---|---|
| 1 | 99 TN | 6 | 99 FN |
| 2 | 98 TN | 7 | 99 FN |
| 3 | 99 TP | 8 | 90 TP |
| 4 | 96 TP | 9 | 94 TP |
| 5 | 99 TN | 10 | 89 TP |

(iii)*Bounding boxes as detected results*

Mask R-CNN algorithm utilizes a fully connected network for mask prediction in the output image. Mask generated around the target object contains spatial information about the object as shown in Fig 7 and 8. While comparing, it is found that instance segmentation is more beneficial as compared to semantic segmentation, to differentiate between samples having the same classes. It generates a colored mask in sample images to represent different objects with different colors, and due to this reason, it is mostly preferred for object detection tasks. As compared to Mask R-CNN, Faster R-CNN takes least inference time and uses region proposal network to predict the region proposals. The predicted region proposals are re-shaped by using RoI pooling layer, which is also used to classify the image within the proposed region and even predicting the offset values for the bounding boxes.

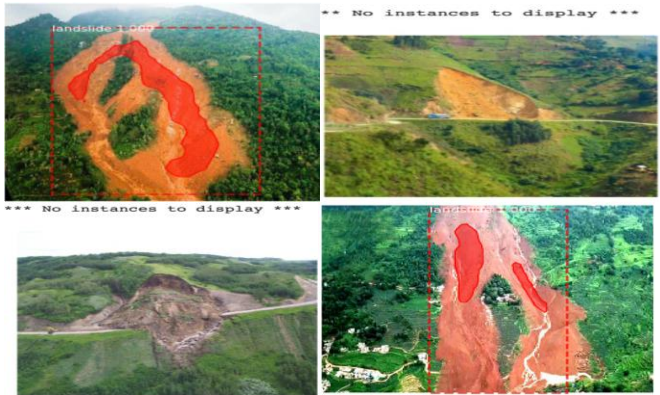

Fig.7.  Landslide detections from ResNet-50

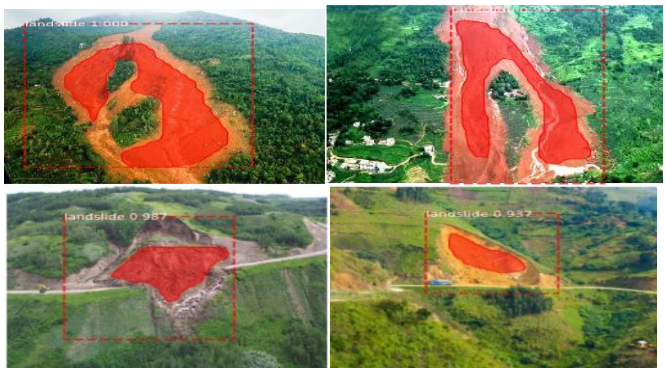

Fig. 8.  Landslide detections from ResNet-101

From the table II and III, it is found that correctly predicted positive observation for landslides is 1 with ResNet-50 for both datasets A and B. While using ResNet-101 correct prediction for the landslide is 1 with dataset A, and 0.93 with dataset B. It is also observed that in ResNet-50 model with dataset A, out of all positive samples, 0.87 landslide samples are picked by the model. Whereas with dataset B correctly chosen samples are 0.73. In the case of ResNet-101, correct landslide samples taken by the model corresponds to 0.93 with dataset A and 0.87 with dataset B. Furthermore, it is observed that landslide classification accuracy is 0.93 when dataset A is used whereas classification accuracy reaches 0.85 when dataset B is used with ResNet-50. When ResNet-101 is used, classification accuracy reaches 0.97 using dataset A, and it reaches 0.9 with dataset B.

## V.  Discussion

Landslide detection by using deep learning frameworks requires a large number of datasets with high resolution, for building the models. Developing a deep learning-based model for landslide detection may be a significantly complex task. In this work, a pre-trained framework, a Mask R-CNN is used to extract landslides of a specific terrain. Transfer learning by using pre-trained deep learning frameworks has been also exploited, by showing a promising approach for improving the object detection performance in deep neural networks. In the proposed research work, the Google Colab has been used for training and, two preconfigured neural networks, i.e. ResNet-50 and 101, have been used as backbone models and for the accuracy assessment of the proposed method. Evaluation was done by using three metrics, i.e. Precision, Recall and F1 Measure. Two different datasets were created: Set A (160 images) and Set B (121 images), for training and testing. To improve the results, an Image Data Generator was applied to artificially expand the size of the training dataset by creating modified versions of the images present in the dataset. Model training was performed by using COCO dataset, and trained weights were utilized for landslide detection, and the output layer (head of the model) was trained by using landslide labelled images. In the process of landslide detection using the proposed methodology, 20 epochs were completed, with unitary batch size.

We could observe that as the number of epochs increased, more number of times the weight has changed in the neural network and the cost function curve goes from under-fitting to optimal and then to overfitting curve. Five object classes were used, i.e. Landslide, Vegetation, Water body, Building and Background. Simulation results revealed that high F1 scores have been reached. The total number of 31 landslides were detected from ResNet-101 using160 images with the highest 1.00 Precision, 0.93 Recall and 0.97 F1 score. From the above results, it is concluded that the proposed methodology can be used for distributed landslide detection, but at the same time, it is still a challenging task to get sufficient amount of training samples with high resolution. Further, we are planning to use an object classification method instead of object detection to improve the accuracy of landslide detection.

In Table VI, we have compared the performance of the proposed method with few important and relevant works existing in literature. It is worth to underline that we could find only one work which was attempting to use transfer learning in combination with Deep Learning [36]. Main comparison has been done with a review paper [37], which has analyzed the performance of many methods applied in similar conditions.

In this latter an analysis was implemented to compare a number of machine learning methods and also assess their performance of classification. The results obtained are inferior in terms of performance to our method, employing a hybrid mask R-CNN. Moreover, while in [37] the study was subjected to some selected areas, the proposed method covers landslide studies and cases of different regions, and it is applicable both to India and Italy in larger perspective (and in general to other different countries). In the other work [36], landslide detection was specifically implemented using deep learning and transfer learning. However, the Confusion matrix has been only used as performance measure and the evaluation was performed for acquisition only. In the proposed work, we have assessed the performance of detection as well as classification with the help of necessary measures and metrics. Moreover, our method has used robust measures for larger geographical contexts. Thus, we consider that the proposed method has outperformed the existing noteworthy research.

TABLE VI
COMPARATIVE ANALYSIS OF THE PROPOSED WORK WITH A FEW WORKS FOUND IN LITERATURE

| Research work | Method | Performance | Classification metrics |
|---|---|---|---|
| [36] | 1. Evaluation of a number of ML methods  2. Applied to certain area for testing | 1. Generalized and average performance | Precision: less than 1.00  Recall: 0.92  F1 Measure: 0.87  Values not obtained for all methods |
| [37] | 1. Deep Learning and Transfer Learning  2. Tested for specific plants and geography | 1. Improved automation of landslide acquisition | Confusion matrix  (Yet, evaluation of detection and classification by other measures are missing) |
| New Mask R-CNN Based Method | 1. Mask R-CNN, a hybrid framework  2. Tested for both Italy and India geographic contexts | 1. Assessed with the help of many metrics as shown from Table I to Table III  2. Highlighted evaluation and detection as well as classification  3. Improved detection of landslide also achieved | Precision: 1.00  Recall: 0.93  F1 Measure: 0.97  Better Robustness |

VI. CONCLUSION

This paper proposed an image-based landslide detection method by combining a pre-trained Mask R-CNN model with transfer learning. In the proposed work, a deep neural network, i.e. a Mask R-CNN, was used to test the performance of the model on landslide images. After getting the results, it is concluded that ResNet-101 performs better than ResNet-50, with 1.00 Precision, 0.93 Recall and 0.97 F1 measures, with a total number of 31 landslide detections on dataset A. Also, landslide datasets used in the paper provide unique data references for a similar type of research. Moreover, the proposed method has outperformed the existing relevant research both in terms of performance measures and robustness. The main problem that we have faced is the annotation of training images which is a very time-consuming task. An object classification methodology instead of the object detection would be attempted in future to improve the accuracy for the landslide detection while resolving the manual annotation of images. This will allow the proposed model to be trained for multiple types of terrains, which is not based on masked instances. Further study also includes improvement in traversal scheme so that, loss value will be monitored.

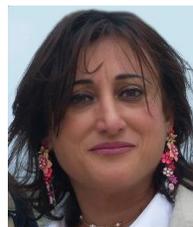

**Silvia Liberata Ullo** graduated with laude in Electronic Engineering at the Faculty of Engineering, of Federico II University, Naples, in 1989. She received the M.Sc. degree from the Massachusetts Institute of Technology (MIT) Sloan Business School of Boston, USA, in June 1992. Since 2004 she is a researcher with the University of Sannio di Benevento, where she teaches Signal theory and elaboration, Telecommunication networks, the Optical and radar remote sensing. She has authored 72 research papers in reputed journals and conferences, and her research interests mainly deal with signal processing, remote sensing, image and satellite data analysis, machine learning applied to satellite data, sensor networks. She is an Industry Liaison for the IEEE Italy Joint ComSoc/VTS Chapter, a member of the Academic Senate at University of Sannio, and the National Referent for the FIDAPA BPW Italy Science and Technology Task Force. Awarded in 1990 with the Marisa Bellisario prize from the homonymous foundation, and with the Marisa Bellisario scholarship from ITALTEL S.p.A company. She has been with ITALTEL, since September 1992, and served as a Chief of some production lines at the Santa Maria Capua Vetere factory (CE), until January 2000. She won a public competition and worked at the Center for Data Processing (CED) in the Municipality of Benevento, from January 2000 to January 2004. In February 2004, she won a researcher contest at the Faculty of Engineering, at University of Sannio, Benevento.

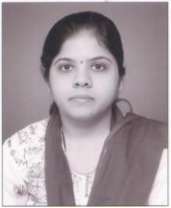

**Amrita Mohan**, currently Pursuing Ph.D. in GIS Cell at Motilal Nehru National Institute of Technology (MNNIT) Allahabad, Prayagraj, UP, India. Her areas of interest include application of Machine Learning in image processing with special reference to detection of landslides, Mobile Ad-Hoc Networks, Wireless Sensor Network.

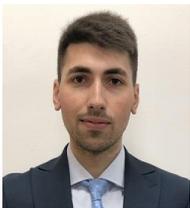

**Alessandro Sebastianelli** graduated with laude in Electronic Engineering for Automation and Telecommunications at the University of Sannio in 2019. He is enrolled in the PhD programme with University of Sannio, and his research topics mainly focus on Remote Sensing and Satellite data analysis, Artificial Intelligence techniques for Earth Observation, and data fusion. He has co-authored some papers to international conferences and submitted two articles under review from important journals for the sector of Remote Sensing. Ha has been a visited researcher at Phi-lab in European Space Research Institute (ESRIN) of the European Space Agency (ESA), in Frascati, and still collaborates with the Phi-lab on topics related to deep learning applied to geohazard assessment, especially for landslides, volcanoes, earthquakes phenomena. He has won an ESA OSIP proposal in August 2020 presented with his PhD Supervisor, Prof. Silvia L. Ullo.

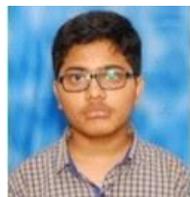

**Shaik Ejaz Ahamed** is a 2020 batch Graduate from Motilal Nehru National Institute of Technology (MNNIT) Allahabad with bachelor's degree in Electronics and Communication Engineering. He was a Research Intern at Defence Research and Development Organisation (DRDO) where he worked on Real Time Operating Systems (RTOS). His area of research includes Machine Learning, NLP, Deep Neural Networks, Distributed Systems, Web Development, System Design, Databases, Image and data processing.

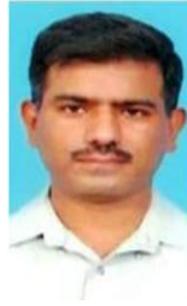

**Basant Kumar**, currently working as Associate Professor in Department of Electronics and Communication Engineering, Motilal Nehru National Institute of Technology, Allahabad, obtained his Ph.D. in Electronics Engineering from Indian Institute of technology, Banaras Hindu University, Varanasi, India (IIT-BHU) in 2011. His area of research includes telemedicine, data compression and medical image processing. He has published more than 70 research papers in reputed international journals and conferences.

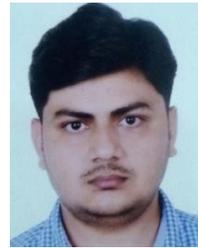

**Ramji Dwivedi**, currently working as Assistant Professor in GIS Cell, Motilal Nehru National Institute of Technology, Allahabad, obtained his Ph.D. in GIS and Remote Sensing from MNNIT Allahabad in 2016. His area of research includes GNSS and InSAR techniques (core and application). He has published more than 15 research papers in reputed international journals and conferences.

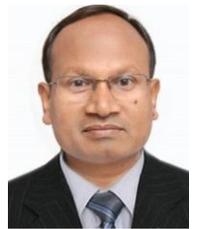

**G. R. SINHA** received the Ph.D. degree. He is an Adjunct Professor with the International Institute of Information Technology Bangalore (IIITB) and currently deputed as a Professor at the Myanmar Institute of Information Technology (MIIT), Mandalay, Myanmar. He is a Visiting Professor (Honorary) with Sri Lanka Technological Campus Colombo for one year, from 2019 to 2020. He has more than 200 research papers, edited books, and books into his credit. He has edited books for reputed International publishers. He has teaching and research experience of 21 years. He has been the Dean of Faculty and an Executive Council Member of CSVTU and currently a member of Senate of MIIT. He has been delivering ACM lectures as a ACM Distinguished Speaker in the field of DSP, since 2017, across the world. His research interests include biometrics, cognitive science, medical image processing, computer vision, outcome based education (OBE), and ICT tools for developing Employability Skills. He is a Fellow of the Institute of Engineers India and a Fellow of IETE, India. He served as a Distinguished IEEE Lecturer in IEEE India council for Bombay section. He was a recipient of many awards and recognitions at national and international level. He has delivered more than 50 Keynote/Invited Talks and Chaired many Technical Sessions in International Conferences across the world. He has eight Ph.D. Scholars, 15 M.Tech. Scholars, and has been Supervising one Ph.D. Scholar. He is active reviewer and editorial member of more than 12 reputed International journals in his research areas, such as IEEE Transactions, Elsevier journals, and Springer journals.